\documentclass[letterpaper, 10 pt, conference]{ieeeconf} 
\IEEEoverridecommandlockouts
\usepackage[dvipdfmx]{graphicx}

\usepackage{epsfig} 

\usepackage{url}
\usepackage{amsmath}
\usepackage{amssymb}
\usepackage{amsfonts}
\usepackage{newtxtext,newtxmath}

\usepackage{algorithmic}
\usepackage{algorithm}

\usepackage{mdframed}
\usepackage{listings}

\usepackage{hyperref}

\title{\LARGE \bf
 {SuctionPrompt}: Visual-assisted Robotic Picking with a Suction Cup Using Vision-Language Models and Facile Hardware Design
 }
\author{Tomohiro Motoda$^{1}$, Takahide Kitamura$^{1}$, Ryo Hanai$^{1}$, Yukiyasu Domae$^{1}$
\thanks{$^{1}$ All authors are with the National Institute of Advanced Industrial Science and Technology (AIST), Tokyo, 135-0064, Japan
}
}

\begin{document}

\maketitle
\thispagestyle{empty}
\pagestyle{empty}

\begin{abstract}
The development of large language models and vision-language models (VLMs) has resulted in the increasing use of robotic systems in various fields. However, the effective integration of these models into real-world robotic tasks is a key challenge. We developed a versatile robotic system called SuctionPrompt that utilizes prompting techniques of VLMs combined with 3D detections to perform product-picking tasks in diverse and dynamic environments. Our method highlights the importance of integrating 3D spatial information with adaptive action planning to enable robots to approach and manipulate objects in novel environments. In the validation experiments, the system accurately selected suction points 75.4\%, and achieved a 65.0\% success rate in picking common items. This study highlights the effectiveness of VLMs in robotic manipulation tasks, even with simple 3D processing. 
\end{abstract}



\section{Introduction}
Large language models (LLMs) and vision-language Models (VLMs) have undergone rapid developments, demonstrating remarkable flexibility in generating relevant outputs, even in novel environments. Their application in visual and language tasks has garnered attention in various fields including robotics, where they are increasingly used to address complex and diverse scenarios. The integration of these models into robotic systems enhances their capacity to handle a broader range of tasks in real-world applications, thereby rendering them more adaptive to dynamic and novel environments. However, despite the promising performance of VLMs, their application is limited in robotic contexts, particularly in recognizing spatial relationships between objects and environments solely through visual inputs. This emphasizes the need to bridge the gap between VLM-generated insights and physical interactions in robotic systems by considering the challenges of spatial recognition and task execution.

\begin{figure}[t]
    \centering
    \includegraphics[width=1.\linewidth]{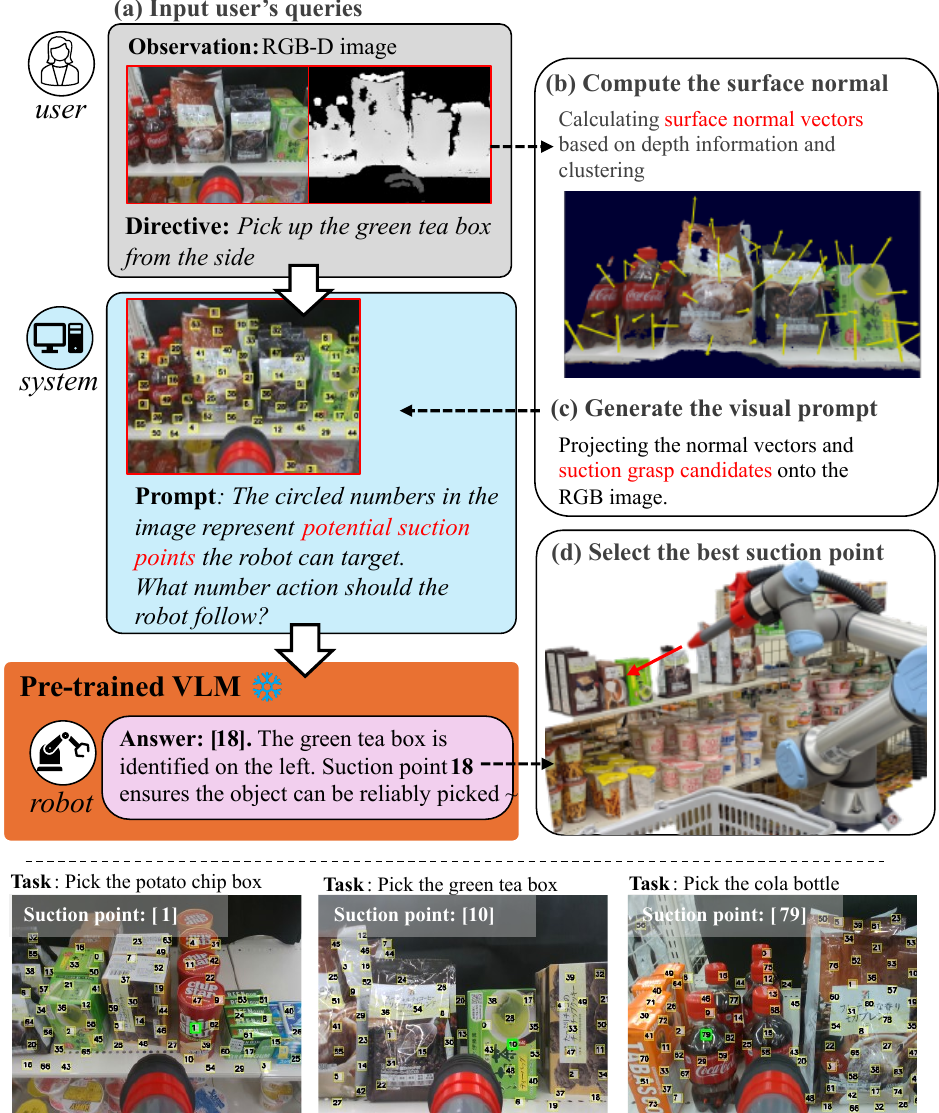}
    \vspace{-20pt}
    \caption{\textbf{Overview of the proposed SuctionPrompt system for robot manipulation tasks}. (a) RGB-depth (RGB-D) image and directive text are input. (b, c) Suction points are generated from estimated 3D surface normal vectors.(d) Robot is instructed to pick up a green-colored tea box by the vision-language model (VLM). The bottom panel shows various object-picking tasks (potato chip box, green tea box, cola bottle, etc.) with corresponding suction points for successful grasping.}
    \label{fig:top}
    \vspace{-10pt}
\end{figure}

In the context of societal demand, the increasing pressure to automate repetitive tasks, such as product handling in convenience stores, has become a significant concern. Convenience stores, which are characterized by a vast range of products (often exceeding 3000 types) and the frequent inclusion of novel objects, present a challenging yet critical environment for robotic automation. Labor shortages, rising operational costs, and the growing demand for faster and more efficient inventory management have led to the development of advanced systems that are capable of performing complex picking tasks. These challenges must be addressed for improving operational efficiency and responding to broader economic and labor market shifts, where automation can fill crucial gaps. Robotic manipulation systems that are capable of handling various objects, including unfamiliar or novel items, can significantly reduce human labor involvement while maintaining accuracy and speed in the restocking and order fulfillment processes.

This paper proposes a versatile robotic manipulation system, called \textbf{SuctionPrompt}, which utilizes a suction-cup-based gripper combined with VLMs. As shown in Fig.~\ref{fig:top}, \textbf{SuctionPrompt} leverages zero-shot object handling, specifically targeting product-picking tasks in convenience stores. By integrating depth information from RGB-depth (RGB-D) cameras, we aimed to provide critical visual support for the robot, ensuring accurate interaction with various objects regardless of their shape or material. Moreover, we introduce a framework for utilizing prompting techniques in VLMs to assist in decision making, enabling the system to perform effectively in complex real-world settings without additional learning from the training data. 
In robotic control, the differences in robot hardware configurations often present challenges owing to gaps in design and implementation. In this study, the relative relationship between the robot manipulator and workspace is partially represented in the images, which affects the performance of VLMs and influence of prompts. To benchmark our system, we built a suction gripper system for comparative evaluation. This paper also discusses the hardware configuration of the system. 

In summary, our primary contributions are as follows: 

\begin{itemize} 
\item A novel prompting method that uses visual cues and text-based commands to guide suction-based robotic manipulation, allowing for effective handling of diverse objects, including previously unseen items. 
\item A three-dimension spatial information-driven action candidate generation method, which enhances the ability of robots to perceive object surfaces and plan actions, improving grasping accuracy and flexibility in unstructured environments. 
\item A hardware and system design that enables zero-shot manipulation by leveraging suction technology and flexible visual prompting, reducing the need for extensive training and precise object positioning in real-world applications. 
\end{itemize}

\begin{figure*}[t]
        \centering
        \includegraphics[width=\linewidth]{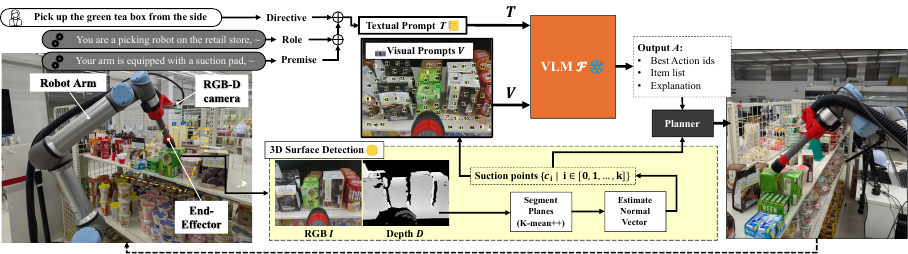}
        \vspace{-15pt}
        \caption{\textbf{Overview of the robotics system with SuctionPrompt}. We propose a versatile robotic manipulation system using a suction-cup-based gripper combined with VLMs to achieve zero-shot object handling, specifically targeting a product-picking task in convenience stores. By integrating depth information from RGB-D cameras, we aim to provide a critical visual prompting for the robot, ensuring accurate interaction with various objects.}
        \label{fig:overview}
\end{figure*}

\section{Related Work}
This section briefly reviews the background and prior studies on VLMs, LLMs, and their integration in robotics. 
  
    \textbf{Large Language and Vision-Language Models}. 
    Both LLMs and VLMs have advanced rapidly~\cite{gpt1,gpt-3,survey1_vision, gemini}. The integration of various modalities, such as language and vision, has proven effective in robotic control, particularly, by exploring the design principles of using LLMs, such as Chat Generative Pre-Trained Transformer (GPT)~\cite{chatGPT}, for robotics. Applications in the real world have been introduced in areas such as embodied agents, navigation~\cite{Navigation-Shah2023, pointing-gesture, pivot}, and planning~\cite{moka, chatGPT-for-robotics}, demonstrating the potential of such models to plan and execute tasks based on high-level commands~\cite{saycan, sayplan, progprompt, code-as-policy, innermonologue, Text2motion}, which enables the generation of robot task plans using LLMs, highlighting the benefits of multimodal integration for enabling more flexible and adaptive robot behavior. 
    These models represent common and practical scenarios and algorithms, and have demonstrated the ability to solve a wide range of tasks in a zero-shot manner. However, to fully harness their advanced capabilities, carefully designed prompts play a crucial role in guiding the models toward effective task execution. 

    \textbf{Visual Prompting for Robotic Manipulation}. 
    Although prompt engineering has been studied extensively in the context of LLMs, researchers are increasingly focusing on visual prompts in VLMs~\cite{Chen2022}. VLMs face unique challenges because of their reliance on visual information, leading to the development of prompt types beyond text input. Shtedritski et al.~\cite{circleclip} discovered that Contrastive Language–Image Pretraining (CLIP)~\cite{clip} can understand a simple circle drawn on an image. CLIPVisual prompts, such as points, markers, and bounding boxes, as seen in the segment anchoring model~\cite{sam}, have been introduced to enhance task performance by incorporating spatial cues alongside textual information. These advancements in prompt engineering, particularly in VLMs, are crucial for functions requiring the integration of visual and linguistic modalities, such as robotic manipulation~\cite{pivot, moka} and scene understanding~\cite{set-of-mark}. Yang et al.~\cite{set-of-mark} demonstrated that set-of-mark prompting can improve visual grounding in GPT 4 with Vision (GPT-4V)~\cite{gpt-4v}, enabling more accurate object localization and interaction. Liu et al.~\cite{3daxiesprompts} proposed the use of 3D axes-based prompts to enhance the ability of the GPT-4V to handle spatial tasks. Building on this, some existing studies have utilized the broad knowledge of LLMs and VLMs for robot manipulation. Nasiriany et al.~\cite{pivot} introduced the concept of iterative visual prompting to enable VLMs to elicit actionable knowledge for robots. Similarly, mark-based visual prompting, as explored by Fang et al.~\cite{moka}, enables robots to perform open-world manipulation tasks, demonstrating the versatility of visual prompts in diverse settings. In robot navigation, Tanada et al.~\cite{pointing-gesture} proposed a visual robot navigation system that interprets human gestures by pointing toward desired directions and moving following Visual Question Answering (VQA). 

    These studies illustrate how straightforward yet innovative concepts can be applied to achieve effective robotic control. In particular, Prompting with Iterative Visual Optimization (PIVOT)~\cite{pivot} has been a significant motivation for our research. It exemplifies how to effectively harness the capabilities of VLMs while linking them to robotic control. However, given the crucial integration of hardware and visual information in robotic systems, careful consideration of hardware requirements~\cite{Fujita2019} and real-world action sets is beneficial when designing prompts. 
    Thus, the proposed method and engineering techniques for leveraging VLMs with real-world robotic models were designed to maximize the potential of VLMs, making them more applicable to a wide range of real-world tasks in the future.

    \textbf{Multimodal Integration for Robotic Control}. 
    Robust and precise perception modules are essential for real-world applications, particularly in robotic manipulation. Tasks in physical environments require models that can accurately process and interpret visual and sensory data because text-based input alone has limitations~\cite{firoozi2024foundation}. This is where research on visual prompting offers valuable capabilities by enhancing the interpretability of tasks that require perception. This provides a more holistic approach to robot decision making and manipulation tasks, leading to a more effective system implementation.

    Recent advances in large-scale models that are specifically designed for robotics, which are often called foundation models, are worthy of attention. A key challenge is that these models must output action control values directly as actionable commands. New model tuning is also required owing to the embodied agents and varying environmental conditions, particularly with the involvement of multiple research communities~\cite{gato, oxe, droid} in these projects to actively develop a general foundation model~\cite{octo, openvla}. However, vast amounts of data are required for training such large-scale models for robotic operation scenarios and for model development. While VLM and prompt-related research may initially appear like a collection of empirical insights, they provide direct access to the inherent zero-shot capabilities of models, allowing for a deeper understanding of model performance. This is particularly significant when these models are applied to specific domains, offering guidelines for their integration into practical applications. 
    Our proposed SuctionPrompt tackles the challenges of zero-shot robotic manipulation and object handling by utilizing VLMs, spatial observational data, and visual prompts to develop an advanced robotic system that is capable of executing complex manipulation tasks, such as the proposed suction-based grasping system.

    \textbf{Suction Gripper for Picking Various Items}
    The accurate picking of unknown objects is an open problem in robotics~\cite{Fujita2019}. The aforementioned studies provide state-of-the-art object perceptions to the robotics community; moreover, in terms of the heavy task of managing object types, these studies achieve reduced costs. However, the process by which a robot system grasps and picks objects is not known, which is a different issue; we can call it the self-embodiment problem in the latest vision models~\cite{embodiedAI}. Prior studies that focused on picking challenges in two competitions: Amazon Picking Challenge and Amazon Robotics Challenge ~\cite{Correll2015, Morrison2018}, identified gripper designs that can enable the picking and placing of various items in warehouses. Notably, several challengers have focused on suction grippers that can pick different types of items~\cite{Bonello2017}. When air is drawn out, the suction gripper grasps objects by creating a negative pressure through the vacuum generated inside the suction cup. Thus, owing to the compressibility of air and flexibility of the suction cup, which is fabricated using materials such as silicone, a wide variety of objects, including flat and curved surfaces and soft materials, can be gripped. 
    
    Using a suction-cup-based design, our proposed system exhibited versatile gripping characteristics that align well with the intended purpose of zero-shot robotic control. Furthermore, because this study selects actions from a discrete set of behaviors, it is well suited for flexibly handling various object shapes and positions without requiring precise positioning.

\section{Method}
We propose a novel visual prompting method called \textbf{SuctionPrompt} that leverages VLMs for emergent robotic control by introducing 3D surface information of the observed scene, which is calculated using depth information as a guide that is specifically tailored for robotic manipulation. This study aimed to achieve zero-shot robotic picking using VLMs without the need for additional training. This approach involves an annotation method that inpaints the perspective images of the robot. 

Given a task description \(T\) in the natural language provided by the operator and an image \(I\) from the wrist perspective of the robot, the method selects a discrete action candidate set \(A\) through suction surface sampling and incorporates a depth image \(D\). 

    \begin{figure*}[t]
                \centering
                \includegraphics[width=\linewidth]{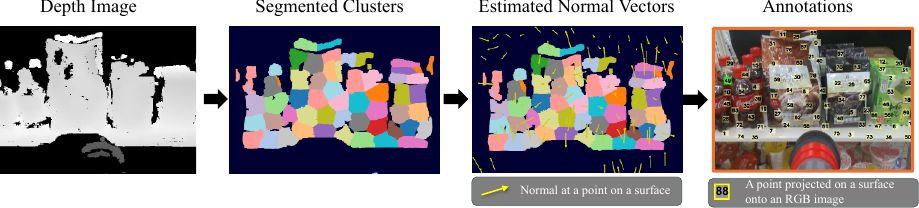}
                \vspace{-20pt}
                \caption{\textbf{Pipeline for a visual prompting}. The process begins by capturing depth images to create 3D point clouds of the scene. These point clouds are then divided into clusters using the K-means++ algorithm. Surface normals are calculated for each cluster, providing important 3D pose information. The 3D points and their corresponding normals are then projected onto the 2D RGB image to create visual cues for candidates on which suction action is to be performed, which are marked with numbered annotations.}
                \label{fig:3d_process}
    \end{figure*}

    \subsection{Integrating Vision-language Models with Robot Control via Visual Question Answering}
    \label{subsec:vqa}
        In this study, we addressed the problem of controlling a robot by detecting suction points from RGB-D images using VQA. Formally, a VLM \(\mathcal{F}\) takes a text sequence \(T\) and an RGB image \(I \in \mathbb{R}^{H \times W \times 3}\) as the inputs, and outputs text sequences \(T'\):
        
        \begin{align}
            T' = \mathcal{F}(I, T)
            \label{eq:vlm}
        \end{align}
        
        In our method, the input text \(T\) consist of a premise description~\(T_{pre}\), role description~\(T_{role}\) for the robot, and directive text~\(T_{dirc}\) from a user, such as, ``Pick up the green tea box from the side,'' as shown in Fig. ~\ref{fig:overview}. These components together form the complete input text $T$, which is structured as follows:
        \begin{align}
            T = \lbrace T_{pre}, T_{role}, T_{dirc} \rbrace. 
        \end{align}
        \(T_{pre}\) provides meta-information that the system requires to generate a response, which helps frame the task within the context of the robot’s operation. For example, ``Your arm is equipped with a suction pad.'' \(T_{role}\) defines the role of the responder, referencing previous method~\cite{kong2023better}, e.g., ``You are a picking robot on the retail store.'' \(T_{dirc}\) represents the explicit directive of users, corresponding to a direct request such as ``Pick up the cola.'', ``Pick the green tea box.'' and `Pick the coffee bag from the side.'' The details of our method are to be mentioned in Sec.~\ref{subsec:vlm}. 
        The VLM is requested to generate a response from the inputs composed of the text \(T\) and image \(I\). A depth image \(D\) is used to detect suction points $S=\{s_1, s_2,...,s_k\}$, which represent the set of all suction points detected by the robot in 3D space. The corresponding projections $o_i = f(s_i)$~(where $i=1,...,k$) are computed using a projection function \(f(\cdot)\). In this study, \(o_i\) corresponds to \(s_i\), and selecting \(o_i\) in the image is implicitly equivalent to selecting \(s_i\). This implies that the visual choice of \(o_i\) in the 2D image plane directly reflects the selection of the corresponding suction point \(s_i\) in 3D space. The function \(\mathcal{V}(I, S)\) adds visual annotations by overlaying suction points $S$ onto the image~\(I\). Here, \(k\) denotes the total number of suction points detected by the robot. We modified Eq.~\ref{eq:vlm} for our approach as follows:
        \begin{align}
            A = \mathcal{F}(\mathcal{V}(I, S), T).
            \label{eq:vp}
        \end{align}
        $A$ represents the textual output generated by the VLM, as shown in Fig. ~\ref{fig:overview}. $A$ can be interpreted as the decision-making response of the system, which informs the robot on how to interact with the detected objects. Moreover, we assumed that $A$ is expressed in a manner that uniquely specifies actions, as is detailed subsequently through prompt adjustments.
    
        To rate the quality of the suction-point detection, we further incorporated additional visual and textual prompts, which are described in the following subsections. 

    \subsection{Three-dimensionally Guided Visual Prompting}
    \label{subsec:vlm}
        As previously mentioned, for the suction-picking task, we aimed to provide visual prompts for the operation of the suction gripper. This study was inspired by PIVOT~\cite{pivot}, which utilized numbered squares on images to indicate potential action candidates. Existing research typically employs random sampling optimization or step-by-step VQA methods for these action candidates. However, because suction actions require precise 6D pose information, determining the necessary gripper pose or orientation using only random points in RGB images is challenging. Although technologies such as 3DAxiesPrompts~\cite{3daxiesprompts} can indirectly infer depth by specifying the axes in images, they must address the challenge of representing irregularly shaped objects. Therefore, we used RGB-D cameras to simultaneously acquire both RGB and depth images, generating 3D point clouds to extract the information required for suction gripping, thereby identifying effective candidate points.

        Fig.~\ref{fig:3d_process} shows the pipeline of the proposed visual prompting model. The process begins by acquiring point-cloud data from RGB-D images. In this study, we reference the suction point sampling method from Dex-Net 3.0~\cite{dexnet3}, adopting candidate points calculated based on geometric processing of 3D information, utilizing surfaces and normal vectors derived from point clouds for suction. 
        These point-cloud data are subsequently segmented using the K-means++ algorithm~\cite{kmeans++}, where the number of clusters is specified as \(k\). The K-means++ algorithm partitions the point-cloud data into \(k\) clusters. Let \(\{C_i\}_{i=1}^k\) represent a set of clusters, where \(C_i\) is the \(i\)-th cluster. For each cluster \(C_i\), surface normals \(\mathbf{n}_i\) are calculated. These normals are determined relative to the centroid \(\mathbf{c}_i\) of each cluster. Let \(\mathbf{p}\) denote a 3D point obtained from the depth data of the RGB-D images. The centroid \(\mathbf{c}_i\) is computed as
        \begin{align}
            \mathbf{c}_i = \frac{1}{|C_i|} \sum_{\mathbf{p} \in C_i} \mathbf{p},
        \end{align}
        where \(|C_i|\) denotes the number of points in the cluster \(C_i\).
        The coordinates of the surface normals \(\mathbf{n}_i\) and centroids \(\mathbf{c}_i\) are then transformed into the RGB image and robot coordinate systems. 
        To project the 3D points onto the 2D RGB image coordinate system, we used the intrinsic parameters of the RGB camera. Let \(\mathbf{K}\) denote the intrinsic matrix of the camera that maps the 3D coordinates to the 2D image coordinates. The projection of the 3D centroid \(\mathbf{c}_i\) onto the 2D image plane is given by
        \begin{align}
            \mathbf{c}_{i, RGB} = \mathbf{K} \cdot \begin{bmatrix}
            \mathbf{c}_i \\
            1
            \end{bmatrix},
        \end{align}
        where \(\mathbf{c}_{i, RGB}\) represents the 2D coordinates of the centroid in the RGB image coordinate system.
        Similarly, to transform the 3D coordinates into the robot coordinate system, we used a transformation matrix. Let \(\mathbf{R}_{robot}\) be the rotation matrix and \(\mathbf{t}_{robot}\) be the translation vector that defines the transformation from the point-cloud coordinate system to the robot coordinate system. The transformed coordinates are as follows:
        \begin{align}
            \mathbf{c}_{i, robot} = \mathbf{R}_{robot} \cdot \mathbf{c}_i + \mathbf{t}_{robot}.
        \end{align}
        
        In the RGB image, markers are overlaid on the calculated coordinates. These markers are represented by numbers enclosed in circles. For each centroid \(\mathbf{c}_{i, RGB}\), a corresponding marker is placed in the RGB image. Simultaneously, the normals and centroid coordinates of the robot coordinate system are preserved.       
        A VLM is employed to make selections based on the overlaid markers. The VLM receives the prompts $T_{role}, T_{pre}, T_{dirc}$ as mentioned in the previous subsection. For example, the role prompt and premise are as explained in Tables~\ref{tab:role},~\ref{tab:premise}, respectively

\begin{table*}[]
\centering
\caption{Assigning roles to vision-language models (VLMs)}
\vspace{-10pt}
\begin{mdframed}
You are a picking robot in a retail store. First, you must detect the objects in the images. If the target object is not found among the detected items, report the issue and return to the initial position. \\
You will be provided with a picking task, and your goal is to pick up the target object safely without dropping it. 
\end{mdframed}
\label{tab:role}
\vspace{-10pt}
\end{table*}

\begin{table*}[]
\centering
\caption{Premises for answering questions based on an image using the VLMs}
\vspace{-10pt}
\begin{mdframed}
    Here is a description of yourself:\\
    - Your arm is equipped with a suction pad, which is visible at the bottom-center part of the image. \\
    - The tip of your arm is directly facing the display shelf. In this scenario, the arm with the black and red suction gripper must be moved so that it becomes perpendicular to the object to be picked. \\
    - Once the arm is in position, initiate the suction to attach and lift the target object.\\
    - Note the following:\\
        Properties of the suction cup:\\
        - The suction cup requires a smooth, flat surface for optimal attachment.\\
        - The picking point should be sufficiently large to accommodate the full suction cup surface.\\
        - The suction cup works best when aligned perpendicularly to the target surface to create a secure seal.\\
        Selection criteria:\\
        - Flatness: Prioritize points on the object where the surface is flat and smooth.\\
        - Accessibility: Ensure that the point is free from obstructions and the arm can easily reach it without collisions.\\
        - Perpendicularity: The arm should be able to align the suction cup perpendicular to the surface at the chosen point.\\
        - Stability: Avoid points near the edges or irregular areas of the object that could lead to an unstable grip.\\
        
    Here is a description of the visualized prompt on the images: \\
    - The circled numbers in the image represent potential suction points that the robot can target. \\
    - Multiple suction points represent good immediate actions to be taken.
\end{mdframed}
\label{tab:premise}
\vspace{-10pt}
\end{table*}

        The VLM evaluates these prompts and selects the most appropriate prompts based on the output format. The tasks of the VLM include:
        \begin{itemize}
            \item Identifying the suction-point numbers.
            \item Recognizing items within the scene.
            \item Providing the rationale and explanation for the selected actions.
        \end{itemize}
        The output from the VLM consists of the chosen suction-point numbers and an explanation of the selections, providing a comprehensive rationale for the decisions made during the picking process.

    \subsection{Suction-Picking Implementation}

    We present an overview of suction picking in Algorithm~\ref {alg:suctionpicking}, and the detailed implementation of each component in this section. The algorithm for the suction-picking system was designed to execute loop sequences for suction-based grasping tasks. As discussed in the previous section, we strategically separated the process into high- and low-level planning, a decision that instilled a sense of confidence in the control of the real robot.  

        \begin{figure}[!t]
            \begin{algorithm}[H]
                \caption{Pseudo-code for Suction Task Execution}
                \label{alg:suctionpicking}
                \begin{algorithmic}[1]
                \STATE Initialize variables and robot pose
                \STATE Load and preprocess camera images (color, depth)
                \STATE Generate initial candidate suction points from point cloud
                
                \WHILE{$count < max\_action\_iterations$}
                    \STATE \text{Obtain current robot pose and images}
                    \STATE \text{Generate candidate points from images (color, depth)}
                    \STATE \text{Process images and point cloud to detect planes}
                    \STATE $A = \mathcal{F}(\mathcal{V}(I, S), T)$
                    
                    \IF{valid target found}
                        \STATE \text{Execute suction action}
                        \IF{vacuum is enabled for pickup}
                            \STATE Wait and perform vacuum-based pickup
                        \ENDIF
                    \ELSE
                        \STATE Reset robot to initial position
                    \ENDIF
                    \STATE $count += 1$
                \ENDWHILE
                \end{algorithmic}
            \end{algorithm}
        \end{figure}

        As outlined in Algorithm~\ref{alg:suctionpicking}, the algorithm commences with the visual prompting techniques that are applied to the object surfaces. This serves as a guiding mechanism for the system, facilitating the generation and subsequent refinement of potential actions. The system operates in an iterative manner, continually assessing the potential points of contact for object manipulation until the task is accomplished or the maximum number of iterations is reached. 
         
        The algorithm captures the prevailing visual input, which includes the color and depth images and current pose of the robot. The generated images provide the initial set of candidate points for performing potential sucking actions, which are derived from the prompts provided to the system. As previously stated~\ref{subsec:vlm}, the function $A_i = \mathcal{F}(\mathcal{V}(I, S), T)$ plays a pivotal role in our algorithm. The loop terminates upon discovering the target, at which point the robot executes its corresponding action. 
        
        A successful pickup process begins with the evaluation of whether the target object is near the tip of the end-effector of the robot. Next, the suction system is activated, and the robot is maneuvered closer to the target object, guided by the surface vector associated with the selected action candidate. Although PIVOT~\cite{pivot} iteratively executes the VQA task to optimize the actions suitable for the picking task, the observation process must also be improved in each step. To gather more detailed scene information iteratively, the robot incrementally approaches the target with its end-effector. If the target is not reached within the specified iteration limit, the robot returns to its initial position to repeat the process.
        Upon a successful approach, with the vacuum enabled, the robot prepares for a suction-based pickup procedure. Once contact is made between the suction cup and target, the predefined motion sequence, as outlined in the experimental design, is executed. This sequence of predefined movements ensures the correct positioning of the suction cup and securing of the object for pickup.        
        The algorithm employs vision-based inputs and visual prompts in an iterative process to refine its grasping actions, thereby providing an adaptable framework for suction-based robotic manipulation tasks. This adaptability ensures that the algorithm can be applied to diverse functions, thereby instilling confidence in its versatility.

\section{Experiments}
    The primary goal was to develop a picking system that can handle a wide variety of items, including previously unseen objects. To validate and analyze the execution of our system, we focused on scenarios in a retail store. 
    \subsection{Experimental Setup}
        \subsubsection{Hardware Setup}  
\begin{figure*}[t]
                \centering
                \includegraphics[width=\linewidth]{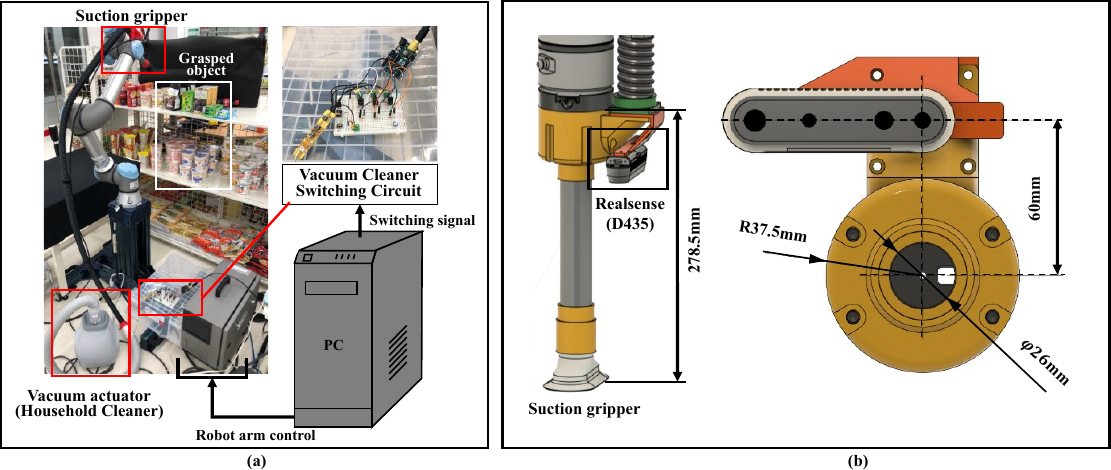}
                \vspace{-20pt}
                \caption{\textbf{Robot arm system for actual machine verification}. (a) Outline of the robot arm system, (b) Suction-gripper-based end-effector.}
                \label{fig:hardware}
            \end{figure*}

            An overview of the proposed system is shown in Fig. ~\ref{fig:hardware}, which consists of a robot arm, suction gripper with an RGB-D camera, and serial controller. The details are as follows. 

            \textbf{Suction Gripper} We employed a suction gripper, which is suitable for grasping objects with different shapes, materials, and surface properties, such as flat or curved surfaces and soft objects. The versatile gripping capabilities of the suction gripper were aligned with the objective of picking diverse items. Moreover, because the system selects actions from a discrete set of behaviors, the ability of the suction gripper to adapt to object shape and position variations without requiring precise alignment makes it particularly advantageous for our use case. However, note that, although this approach offers flexibility, the design of low-level motion planning must be handled separately. Selecting actions from predefined sets imposes certain limitations compared to more flexible text-based decision-making systems.
    
            To generate negative suction pressure, we used a commercially available vacuum cleaner (Panasonic Inc., model number) as a motor-driven vacuum pump. To control the ON/OFF operation of the vacuum cleaner in our system via signal control, we incorporated a metal-oxide-semiconductor field-effect transistor (K2232, Toshiba) into the built-in power-switching circuit of the vacuum cleaner. We assembled a custom-switching circuit using Arduino. Based on the output of the VLM, which determines whether suction is feasible, ON/OFF signals are sent via serial communication to the Arduino, enabling circuit switching.
    
            \textbf{Camera Setup} We used a RealSense D435i RGB-D camera. In our setup, the camera was strategically fixed to the wrist of the robot to ensure a clear view of the immediate working area. Because our system primarily uses three-dimensional spatial information, the position of the camera relative to the robot arm was not critically affected by the mounting position of the camera, provided that the necessary coordinate transformations could be computed. 
            For simplicity, the camera was aligned such that the center of the image coincided with the end-effector of the robot. 
            This alignment ensured that the camera view and robot movements are coordinated in a straightforward manner during the picking tasks. 

        \subsubsection{Software Setup} 
            The software for performing a series of prompts and VQA tasks in our model was implemented based on the Application Programming Interface (API) provided by OpenAI, utilizing the latest version as of 2024, that is, GPT-4o~\cite{gpt-4o}. Although prompts are expected to provide examples or specify formats to ensure that the model maintains an appropriate response format, the accuracy or flexibility of the response primarily depends on the performance of the API, making it unstable at times. Therefore, we adopted JSON Schema offered by the API of OpenAI \footnote{https://openai.com/index/introducing-structured-outputs-in-the-api/}, which is a feature that allows for strict output formatting. This function can currently only be used on the GPT-4o and GPT-4o-mini~\cite{gpt-4o-mini} versions. Note that, in the subsequently presented evaluations, errors or detection failures due to the response format were not considered a fundamental failure of the model and were excluded from the scope of this study. 

    \begin{figure*}[t]
        \centering
        \includegraphics[width=\linewidth]{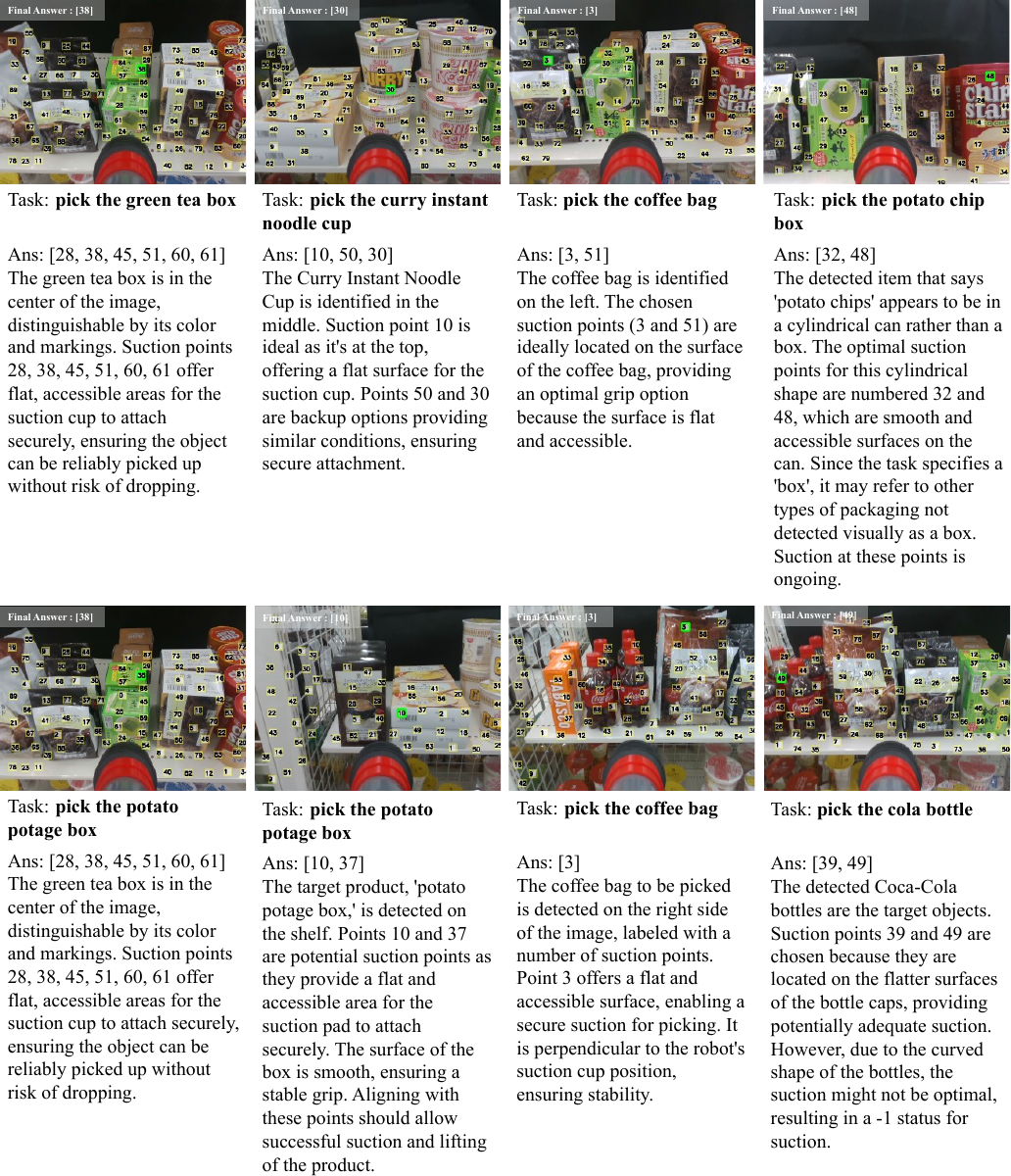}
        \vspace{-20pt}
        \caption{Responses from the vision-language model (GPT-4o) regarding the numbering of suction points on the prompted image, along with the rationale for each selected point.}
        \label{fig:response}
    \end{figure*}

    \subsection{Offline Performance and Ablations}
    \label{subsec:software_test}
    In this section, we evaluate the performance of the entire system, including SuctionPrompt, through offline assessments. As mentioned in previous sections, we performed demonstrations using GPT-4o and GPT-4o-mini to verify whether the proposed method, which involves prompts (including images), correctly addressed the VQA task. For this evaluation, we prepared 138 random test images of the shelves, including depth images. Of these, 114 images were used for validation, excluding those in which GPT-4 could not correctly recognize the object names. These images were pre-annotated using numerical labels. Rather than considering the ability of the robot to pick up objects, we constructed the ground truth based on the subjective judgments of human participants. The accuracy of the system output was validated. The results are shown in Fig.~\ref{fig:response}. 

    \begin{table}[t]
        \caption{Performance comparison between our system utilizing different VLMs on the Visual Question Answering task for object selection from shelves.}
        \centering
        \includegraphics[width=0.8\linewidth]{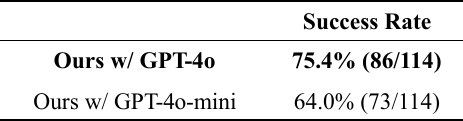}
        \vspace{-10pt}
        \label{tab:response}
    \end{table}

    The results from the GPT-4o and GPT-4o-mini models show that our system can effectively process images of shelves and provide accurate responses for object selection using VQA. As shown in Fig.~\ref{fig:top}, when integrated with GPT-4o, our system achieved an accuracy of 75.4\% (86 out of 114 images), whereas the GPT-4o-mini variant achieved a lower accuracy of 64.0\% (73 out of 114 images). These results demonstrate the advantage of using a more sophisticated model such as GPT-4o for complex scene-understanding tasks. Although both models used the same image-based prompts, the larger GPT-4o model showed superior proficiency in accurately identifying the correct objects in cluttered shelf environments. The results are summarized in Table~\ref{tab:response}. 

    \begin{figure}[t]
        \centering
        \includegraphics[width=\linewidth]{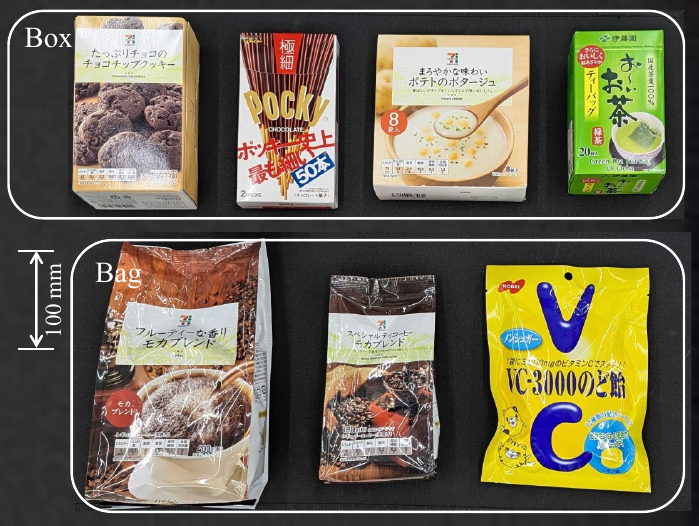}
        \vspace{-20pt}
        \caption{Targeting objects with bag and box shapes for picking tasks. These shapes are commonly found in retail environments, representing a wide range of products that are typically encountered in convenience stores.}
        \label{fig:targetobject}
    \end{figure}

    \begin{figure*}[t]
        \centering
        \includegraphics[width=\linewidth]{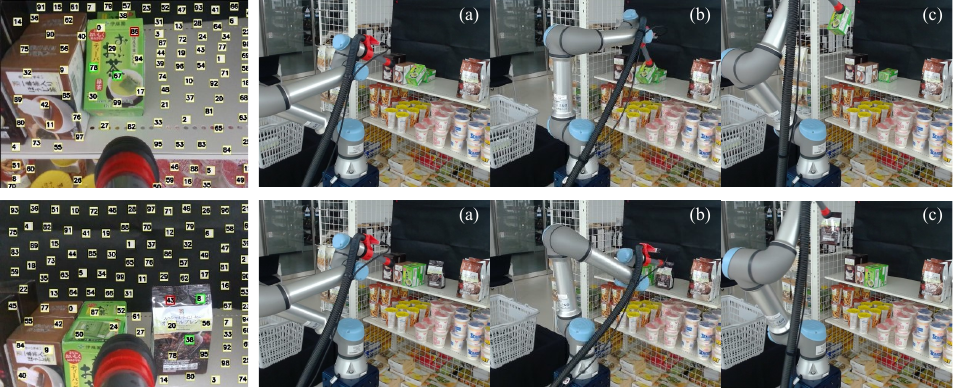}
        \vspace{-20pt}
        \caption{Examples of visual prompts and corresponding robot motions. In each row, the image on the left shows the suction candidates (marked in yellow), better points (marked in green), and the best point (marked in red). The three images on the right (a, b, c) depict keyframes of the motion throughout the grasping trajectory.}
        \label{fig:motion}
    \end{figure*}

    \subsection{Physical Experiments}
    Real-world experiments were conducted to evaluate the proposed prompts and hardware. These experiments were designed to confirm whether the robot can pick up various items from scenes that were not viewed previously. We assumed that the robot received a directive message (equivalent to $T_{dirc}$, as described in Sec.~\ref{subsec:vqa}), which simulated an order from a human customer, and then attempted to pick up an item from a convenience store shelf. The objects were categorized into two types: boxes and bags as shown in Fig.~\ref{fig:targetobject}. The results are summarized in Table~\ref{tab:physical_eperiment}.

    \begin{table}[t]
        \caption{Success rates of the proposed system in real-world experiments using VLM for different object types. }
        \centering
        \includegraphics[width=0.8\linewidth]{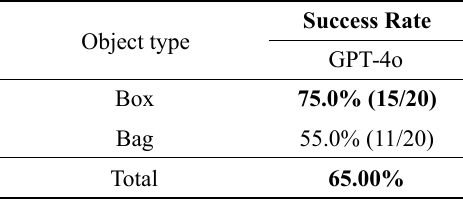}
        \vspace{-10pt}
        \label{tab:physical_eperiment}
    \end{table}
    
    In Fig.~\ref{fig:motion}, we provide two examples of generated motions.
    The robot successfully picked up 75.0\% (15/20) of the box-shaped items and 55.0\% (11/20) of the bag-shaped items, resulting in an overall success rate of 65.0\%. 

    The real-world experiments provided valuable insights into the performance of VLMs in robotic manipulation tasks. The experiments showed that the proposed prompting system and the suction-based hardware effectively enabled the robot to handle various unseen objects from a convenience store shelf. The directive text, which represents human-like instructions, was crucial in guiding the robot to perform the correct actions.

    However, the success rate varied between different object categories, specifically boxes and bags. The VLM-based system performed better with box-shaped items due to their geometric structure, which allowed for more straightforward surface recognition and suction point determination. In contrast, bags presented more challenges because their flexible and irregular surfaces made it difficult for the VLM to generate reliable action candidates based purely on 2D visual input. This highlights a limitation of current VLMs in dealing with deformable objects, where additional 3D spatial reasoning or more advanced perception techniques may be required to assess suitable grasping points accurately.
    
\section{Limitations}
This study aimed to enhance state-of-the-art models by evaluating their performance in real-world VQA tasks using the latest VLM APIs. Existing models rely solely on VLMs to generate valid answers without incorporating robot-specific information through pretraining. This limitation poses challenges in picking tasks, which differ from traditional object recognition tasks. These challenges provide valuable insights for the future applications of models in real-world scenarios.

\textbf{Scene Recognition}
Our system included explicit 3D scene recognition that was designed for robotic manipulation; however, its heavy reliance on VLMs may present a performance bottleneck. As discussed in Sec.~\ref{subsec:software_test}, this issue is evident when comparing GPT-4o with its lighter version, GPT-4o-mini. Recognition becomes impossible if the target object is not visible, and a significant dependence on the image resolution and quality is observed. Notably, the system sometimes struggles to distinguish between overlapping objects when evaluating items such as bags. Although labeling of text and symbols can be helpful for providing guidance, in certain cases, they may hinder scene recognition. For example, visual prompts applied to text may not always be suitable for the tasks at hand.

\textbf{Application to Special Instruction Sets}
This study primarily focused on suction-based actions, expecting the output to include prompts related to the desired suction type or details of the hardware of the robot. However, a fixed camera on the end-effector cannot capture a complete view of the suction gripper, resulting in the loss of important information. Consequently, the specific characteristics of the suction cup are not effectively reflected in the actions of the system. Instead of fixing the camera on the end-effector, incorporating an external camera may improve the performance. Ultimately, the existing knowledge gaps in VLMs remain a bottleneck. Training with robot-specific data or depth images is crucial for addressing these limitations. In the future, approaches such as in-context learning may be pivotal, particularly for sequential tasks that require recalling past experiences. This form of memory-based learning facilitates tasks that require dynamic video-like recognition.

\section{Conclusion}
Using a suction-based grasping approach, this study developed a versatile robotic manipulation system that can handle various items, including objects that have not been previously encountered by the system. Using a suction gripper, we demonstrated that the system can manipulate objects with different shapes, materials, and surface properties, making it suitable for complex real-world picking tasks. Our system employs visual prompts and VLM-based guidance to generate robust actions without requiring extensive training or precise pre-calibration, thereby enabling zero-shot manipulation.

Our method emphasizes the importance of combining 3D spatial information with adaptive action planning, allowing the robot to flexibly approach and manipulate objects, even when their exact positions are unknown. Through validation, this study explored the potential of VLMs for robotic manipulation applications and suggested that even simple 3D processing can be effective in such contexts. Although certain limitations were observed owing to the discrete action sets, our approach strikes a practical balance between adaptability and system complexity, making it effective for tasks in unstructured environments.

\section*{Acknowledgements}
This research was conducted with financial support and using the experimental facilities provided by the National Institute of Advanced Industrial Science and Technology. We express our gratitude for the significant support and assistance provided throughout this study. 
We thank Drs. Natsuki Yamanobe and Abdullah Mustafa for their technical contributions and participation in discussions, which have greatly supported this research. 
We would like to thank Editage (www.editage.jp) for English language editing.

\end{document}